\documentclass[letterpaper, 10 pt, journal, twoside]{ieeetran}
\usepackage{cite}
\usepackage{amsmath,amssymb,amsfonts}
\usepackage{bm}
\usepackage{graphicx}
\usepackage{textcomp}
\usepackage{xcolor}
\usepackage{algorithm}% http://ctan.org/pkg/algorithms
\usepackage{algcompatible}% http://ctan.org/pkg/algorithmicx
\usepackage{algpseudocode}
\usepackage[english]{babel}
\usepackage{booktabs}
\usepackage{hyperref}
\usepackage{cleveref}[capitalise] %for Cref
\usepackage{float}
\newtheorem{theorem}{\textbf{Theorem}}
\usepackage{adjustbox}
\usepackage{subcaption}
\usepackage{enumitem}
\setlist[enumerate]{topsep=0pt, partopsep=0pt, itemsep=0pt, parsep=0pt}
\setlist[enumerate]{leftmargin=*}

\usepackage{etoolbox}
\makeatletter
\patchcmd{\@makecaption}
  {\scshape}
  {}
  {}
  {}
\makeatletter
\patchcmd{\@makecaption}
  {\\}
  {:\ }
  {}
  {}
\makeatother
\usepackage[skip=0pt,font=small,labelfont=bf]{caption}

\newcommand{\bb}{\mathbf}  
\newcommand{\blue}[1]{{#1}}
\newcommand{\rebut}[1]{{#1}}
\algnewcommand\RETURN{\State \algorithmicreturn}%

\begin{document}
\bstctlcite{IEEEexample:BSTcontrol}

\author{
Herbert Wright$^{1}$, Weiming Zhi$^{2,3}$, Martin Matak$^{1}$, Matthew Johnson-Roberson$^{2,4}$, Tucker Hermans$^{1,5}$
\thanks{Manuscript received: June, 30, 2025; Revised September, 24, 2025; Accepted October, 15, 2025.}
\thanks{This paper was recommended for publication by Editor Markus Vincze upon evaluation of the Associate Editor and Reviewers' comments.
This work was supported in part by NSF Award \#2321852.}
\thanks{$^{1}$Herbert Wright, Martin Matak, and Tucker Hermans are with the University of Utah Robotics Center and Kahlert School of Computing, University of Utah, Salt Lake City, UT, USA}
\thanks{$^{2}$Weiming Zhi and Matthew Johnson-Roberson are with the College of Connected Computing, Vanderbilt, TN, USA}
\thanks{$^{3}$Weiming Zhi is with the School of Computer Science, The University of Sydney, Australia}
\thanks{$^{4}$Matthew Johnson-Roberson is with the Robotics Institute, Carnegie Mellon University, Pittsburgh, PA, USA}
\thanks{$^{5}$Tucker Hermans is with NVIDIA Corporation, Seattle, WA, USA}%
\thanks{Digital Object Identifier (DOI): see top of this page.}}

\title{
    Robust Bayesian Scene Reconstruction with Retrieval-Augmented Priors for Precise Grasping and Planning
}

\frenchspacing
\maketitle

\begin{abstract}
Constructing 3D representations of object geometry is critical for many robotics tasks, particularly manipulation problems. These representations must be built from potentially noisy partial observations. In this work, we focus on the problem of reconstructing a multi-object scene from a single RGBD image using a fixed camera. Traditional scene representation methods generally cannot infer the geometry of unobserved regions of the objects in the image. Attempts have been made to leverage deep learning to train on a dataset of known objects and representations, and then generalize to new observations. However, this can be brittle to noisy real-world observations and objects not contained in the dataset, and do not provide well-calibrated reconstruction confidences. We propose BRRP, a reconstruction method that leverages preexisting mesh datasets to build an informative prior during robust probabilistic reconstruction. We introduce the concept of a retrieval-augmented prior, where we retrieve relevant components of our prior distribution from a database of objects during inference. The resulting prior enables estimation of the geometry of occluded portions of the in-scene objects. Our method produces a distribution over object shape that can be used for reconstruction and measuring uncertainty. We evaluate our method in both simulated scenes and in the real world. We demonstrate the robustness of our method against deep learning-only approaches while being more accurate than a method without an informative prior. Through real-world experiments, we particularly highlight the capability of BRRP to enable successful dexterous manipulation in clutter.
\end{abstract}


\section{Introduction}

\begin{figure}[t]
    \centering
    % \begin{tabular}{cc}
    %     % \adjustbox{valign=m}{\rotatebox[origin=c]{90}{\parbox{3cm}{\centering\small Unwanted\\Collision}}} &
    %     \includegraphics[width=0.4\columnwidth]{fig2_1-1.png} &
    %     \includegraphics[width=0.4\columnwidth]{fig2_1-2.png} \\
        
    %     % \adjustbox{valign=m}{\rotatebox[origin=c]{90}{\parbox{3cm}{\centering\small Precise\\Collision\\Avoidance}}} &
    %     \includegraphics[width=0.4\columnwidth]{fig2_2-1.png} &
    %     \includegraphics[width=0.4\columnwidth]{fig2_2-2.png}
    % \end{tabular}
    \includegraphics[width=0.9\linewidth]{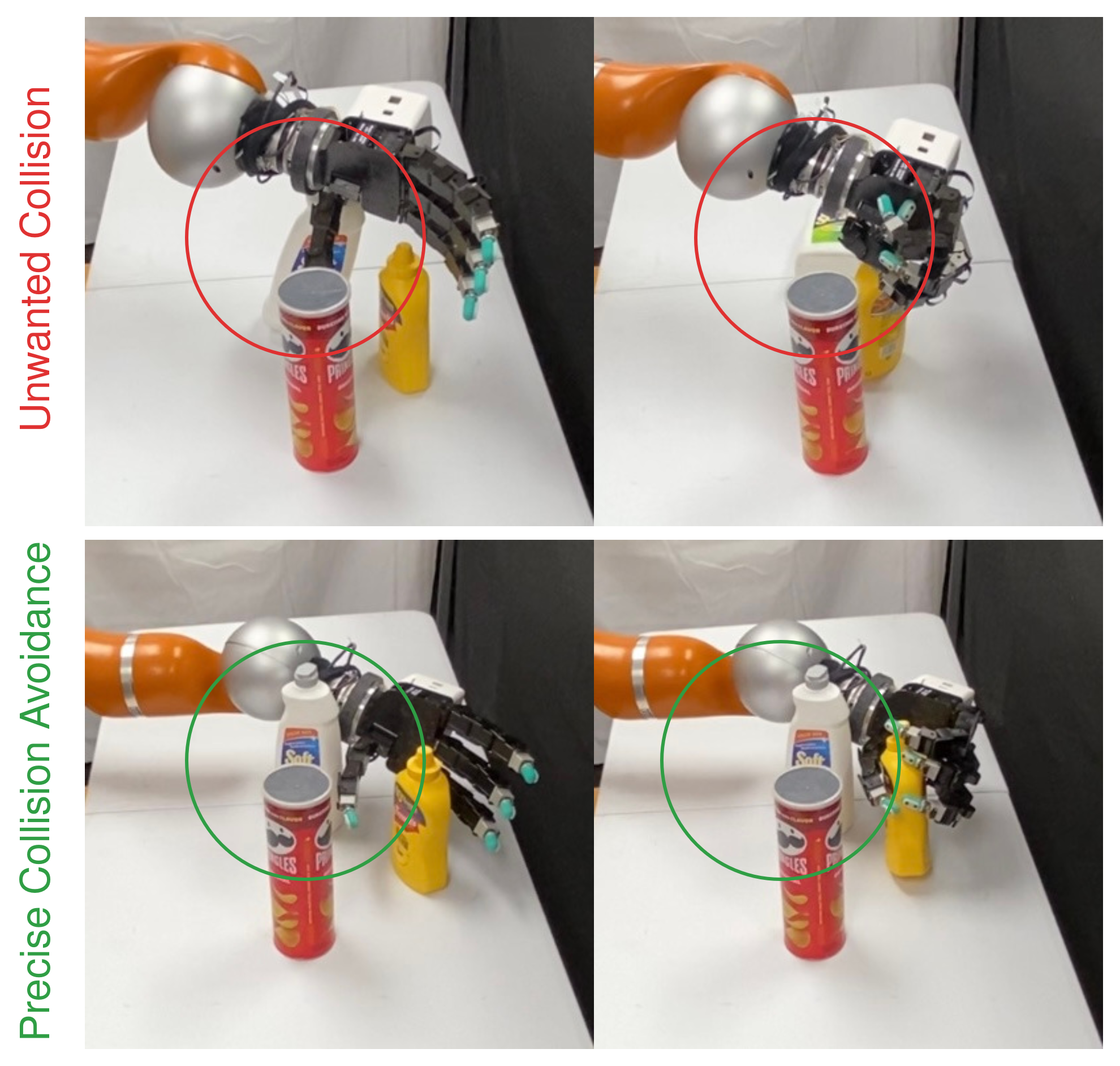}
    \caption{Poor understanding of geometry can lead to unwanted collisions during robotic grasping (\textbf{top}), whereas robust, accurate reconstructions can enable precise collision avoidance during manipulation (\textbf{bottom}).}
    \label{fig:importance-of-geometry}
\end{figure}

\IEEEPARstart{T}{o} operate autonomously in novel settings, a robot must construct rich internal representations of its environment. These representations need to be particularly fine-grained for robotic manipulation, which often requires closely interacting with and avoiding objects. \blue{Consider the task of robotic grasping in clutter; a robot needs to avoid unwanted collisions, while also operating in close proximity to one or more objects to make the necessary contact for the grasp. State of the art grasp synthesis methods often rely on motion planners to reach grasp poses in clutter~\cite{chen2024springgrasp,de2024task,matak2022planning,martin-arxiv-fpte}, which need accurate 3D geometry for collision checking. Thus, a lack of geometric understanding can lead to unwanted collisions causing task failure or worse damage to the robot or environment.} This is pictured in \Cref{fig:importance-of-geometry}. 
Many grasping and motion planning algorithms require explicit 3D representations of the scene's geometry.
These representations must be built from observations that are both noisy and, due to occlusion, only contain partial information of the scene. In our case, we focus on the problem of robustly building a 3D representation of multi-object scenes from a single RGBD camera image.

One approach to this problem is to train a neural network to \textit{predict} the full geometry of an object given a partial view. Such approaches are able to use existing mesh datasets to more accurately infer the occluded backside of objects. Unfortunately, these approaches also tend to have a number of problems when used on real-world depth cameras. The presence of unknown objects, significant occlusion, noisy point clouds, or inaccurate segmentation prevents these methods from being reliably deployed in less-structured environments. 

Deep learning methods for 3D reconstruction generally lack a sense of uncertainty about the shape of the object. This can be particularly detrimental when an object in the scene is only partially observed and we seek to factor in the geometry of observed and occluded regions. Such uncertainty can enable safer and more robust operation in a range of downstream tasks, such as robot grasping \cite{chen2024springgrasp, de2024task}, safe motion generation and active learning.

Another common approach to building full 3D representations from a partial view is to build a representation solely from the observed data without considering prior information from mesh datasets. A common example of this is the Gaussian Process Implicit Surface \cite{dragiev2011gaussian} model. A more recent example is V-PRISM \cite{wright2024vprism}, which probabilistically maps tabletop scenes without using prior information. These reconstruction methods are more robust to unknown objects because they do not rely on any training distribution. However, this also means that they cannot reconstruct the unobserved backside of objects. Humans have the remarkable ability to infer the geometry of scenes based on prior experience, we seek to imbue robots with the same capability. 

\begin{figure}
    \centering
    \includegraphics[width=1\linewidth]{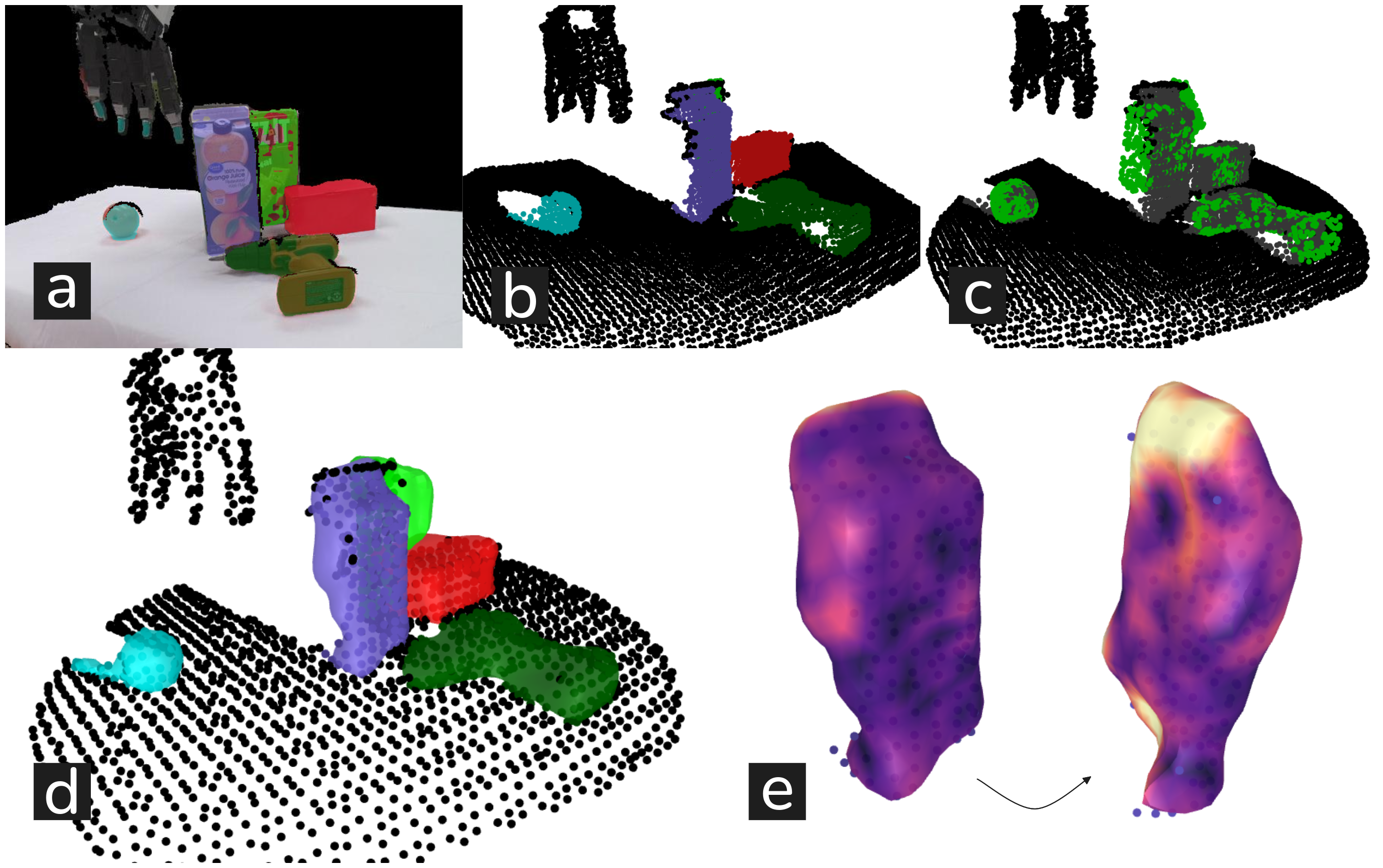}
    \caption{\rebut{BRRP (a), (b) takes an input segmented RGBD image, and (c) \textit{retrieves} objects to use as a prior to (d) reconstruct the scene as well as (e) capture principled \textit{uncertainty} about object shape, in this case, an out-of-distribution orange juice carton}. }
    \label{fig:fig1}
\end{figure}

In this work, we introduce a novel Bayesian approach for robustly reconstructing multi-object tabletop scenes by leveraging object-level shape priors. We present \textbf{B}ayesian \textbf{R}econstruction with \textbf{R}etrieval-augmented \textbf{P}riors  (BRRP). BRRP is resilient to many of the pitfalls of learning-based methods while still being able to leverage an \textit{informative prior} to more accurately reconstruct known objects. To further improve efficiency, we introduce the idea of a retrieval-augmented prior, where we retrieve relevant components of our prior distribution based on classification results. We begin with an observed RGBD image with corresponding instance segmentations. Then, we compute an identification result that predicts which objects from our database should be retrieved to use as a prior during reconstruction. We use this prior along with a sampled likelihood to infer a posterior distribution over object shapes. Because we solve for a distribution, we can recover principled uncertainty about each object's shape. In practice, we use a pre-existing foundation model to perform the identification and use registration to make our prior pose invariant. 
\Cref{fig:fig1} shows an example scene reconstruction, \rebut{where one of the objects (the orange juice) is decidedly out-of-distribution and our method is still able to reconstruct it according to the observed point cloud.}  

We conduct experiments on BRRP in both procedurally generated scenes and in the real world. We quantitatively show that BRRP results in accurate reconstructions and that BRRP is robust to unknown objects in the generated scenes. We qualitatively show that BRRP is robust to noisy real world scenes collected from an RGBD camera, and is able to capture uncertainty within the reconstructions. We also demonstrate downstream improvement with real-world grasping experiments, where the robot is able to more robustly arrive at an accurate pre-grasp pose.

Our three primary contributions can be summarized as:
\begin{enumerate}
    \item The formulation of retrieval-augmented priors for Bayesian inference.
    \item A formulation for a prior over Hilbert maps with pose and scale invariance
    \item A novel, robust \textit{Bayesian} scene reconstruction method that utilizes prior information from existing mesh datasets
\end{enumerate}

We organize our paper as follows. We overview related works in \Cref{sec:related}. In \Cref{sec:background}, we cover mathematical preliminaries for our method. \Cref{sec:priors} introduces retrieval-augmented priors. BRRP is introduced in \Cref{sec:BRRP}. Experiments are found in \Cref{sec:experiments} and a conclusion in \Cref{sec:conclusion}

\section{Related Works}\label{sec:related}

{\bf 3D Representations.} There are many different ways of representing 3D geometry of a scene. In the mapping literature, memory-intensive voxels are used as a representation of the environment. Hilbert maps \cite{ramos2016hilbert}, on the other hand, are a continuous occupancy map that takes the form of a linear function over some hinge point feature space. Hilbert map representations have also been extended to Bayesian Hilbert maps of various forms \cite{senanayake2017bayesian, senanayake2018automorphing, zhi2019continuous}. Neural implicit functions have also been used to represent continuous 3D geometry \cite{park2019deepsdf, mescheder2019occupancy, sitzmann2020implicit}.
Other representations are built using differentiable rendering and combining multiple views \cite{mildenhall2021nerf, kerbl20233d}. Foundation models have also been developed for this task in \cite{Wang_2024_CVPR}  and applied to robotics \cite{sim_geo_pose2024}. We instead focus on the harder problem of reconstruction from only a single-view. 
% Other representation primitives have been studied, including super quadrics \cite{liu2022robust}.

{\bf 3D Reconstruction with Deep Learning Priors.} Many methods have been proposed as ways to leverage deep learning to reconstruct scenes or objects. While some methods aim to predict object shape from RGB data only \cite{xu2024instantmesh, liu2024one, engelmann2021points}, we instead focus on using depth measurements during reconstruction. DeepSDF \cite{park2019deepsdf} is a method to reconstruct an object by running inference-time optimization to recover a latent code for a neural implicit function. In the context of robotics, \cite{liao2024uncertainty} extends DeepSDF to have uncertainty-awareness. Other work, such as occupancy networks \cite{mescheder2019occupancy} or PointSDF \cite{van2020learning} try to directly predict such a latent code without inference-time optimization. Deep learning has also been leveraged to learn kernels, which are used to construct a continuous signed distance function \cite{williams2022neural, huang2023neural}. Language is also used during reconstruction in \cite{kasten2024point} and \cite{cheng2023sdfusion}. Another work uses a voxel-to-voxel variational autoencoder conditioned on bounding boxes \cite{saund2021diverse}. While these works typically focus on single object scenes, other work focuses on reconstructing multi-object scenes. For example, \cite{agnew2021amodal} learns to reconstruct a voxel representation, leveraging different input channels to account for occlusion. Alternatively, work such as \cite{iwase2024zero} aim to reconstruct the scene-wide geometry without bothering to separate each object. Instead, we focus on reconstructing \textit{individual} objects in a manner robust to occlusion. In practice, these deep learning approaches can struggle to reconstruct noisy scenes with multiple, highly occluded, unknown objects on real world depth cameras. 
These methods can also struggle when segmentation is inaccurate.
% Inaccurate segmentation can also be a problem for many of these methods.

{\bf 3D Reconstruction without Deep Learning Priors.} There are also many approaches to perform probabilistic 3D reconstruction without deep learning. Some of such methods for reconstruction use informative prior information by assuming fixed classes of objects, such as 3DP3 \cite{gothoskar20213dp3}. Other methods use an uninformative prior, such as Gaussian process implicit surfaces (GPIS) \cite{dragiev2011gaussian}. While there is an extension of GPIS to a slightly more informative prior \cite{martens2016geometric}, the only priors that can be enforced are specifically spherical, ellipsoidal, cylidrical, or planar priors. We instead derive our prior from pre-existing mesh data. V-PRISM \cite{wright2024vprism} is another method that probabilistically \textit{maps} the scene using a multi-class framing.

{\bf Using Reconstructions in Manipulation.} 3D reconstruction methods have seen extensive use in manipulation. In \cite{van2020learning}, PointSDF provides collision constraints during grasping. PointSDF is also utilized in \cite{matak2022planning}, where tactile sensors are used along with the reconstruction during grasping. A learning-based voxel representation is used for grasping in \cite{lundell2019robust}. Neural shape completion is also used during the anthropomorphic grasping pipeline proposed in \cite{hidalgo2023anthropomorphic}. GPIS is also a common representation for manipulation applications. Some recent work has utilized the uncertainty from GPIS representations during grasp selection \cite{chen2024springgrasp, de2024task}. We believe BRRP provides principled uncertainty measurements that can similarly be utilized in downstream manipulation tasks.

% \vspace{-0.25cm}
\section{Background}\label{sec:background}
% \vspace{-0.25cm}

\subsection{Hilbert Maps}
A Hilbert map \cite{ramos2016hilbert} is a \emph{continuous} occupancy map. It represents the environment by a continuous function that is defined by a linear function of a fixed feature transform. Typically this feature transform is induced by a set of \textit{hinge points}, $\{\bb h_1, ..., \bb h_H\} \subset \mathbb R^3$ and a translation-invariant kernel $k(d)$. The transform is then defined as:
$$ \phi(\bb x) = [1, k(\bb x - \bb h_1), ..., k(\bb x - \bb h_H)]^\top. $$
Typically, a Gaussian kernel is used and hinge points are placed in an evenly-spaced grid. An occupancy map can then be defined by a single weight vector, $\bb w \in \mathbb R^{H + 1}$, as such:
$$ m(\bb x) = \sigma(\bb w^\top \phi(\bb x)). $$

\begin{figure}
    \centering
    \includegraphics[width=1\linewidth]{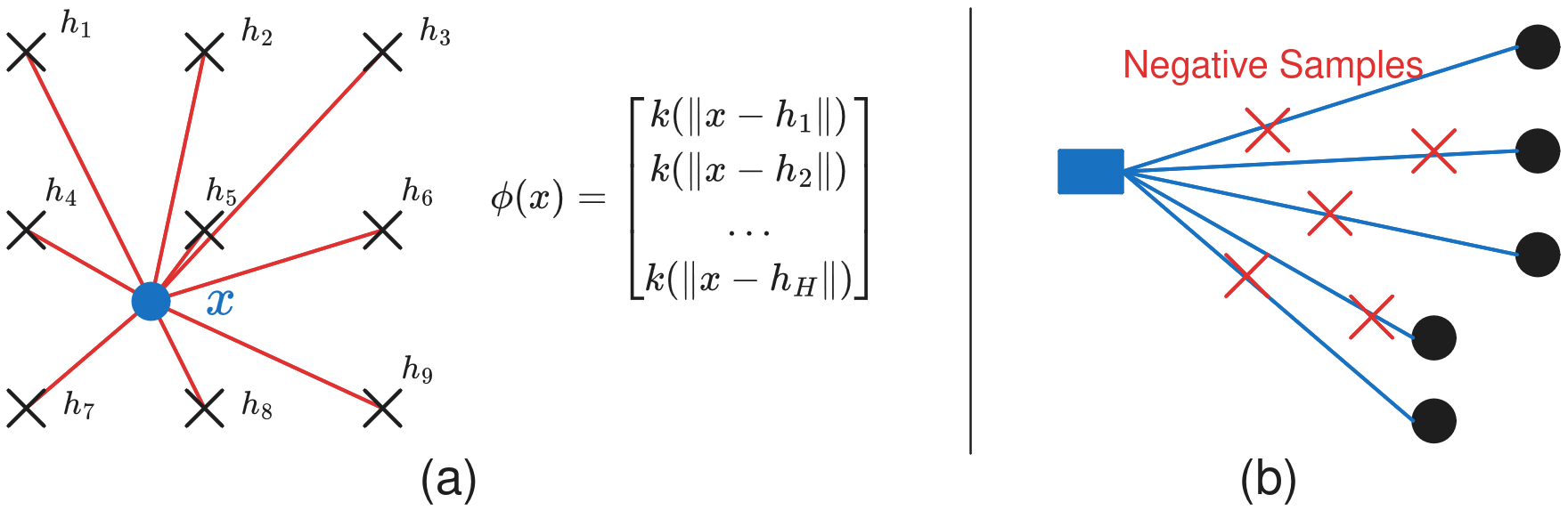}
    \caption{(a) A \textit{hinge point} feature transform induced by a set of hinge points is used by Hilbert maps \cite{ramos2016hilbert}; (b) these maps are built by first sampling \textit{negative samples} along the unoccupied portions of the camera ray.}
    \label{fig:hilbertmaps}
\end{figure}

To recover the weight vector corresponding to a given depth observation, \textit{negative sampling} is performed along the unoccupied portions of the depth rays. These negative samples are assigned a label of unoccupied and the points at the end of the ray are labeled as occupied. Then, stochastic gradient descent (SGD) is performed on the binary cross entropy (BCE) of the negative samples and terminal points of the ray. The binary cross entropy measures the \textit{likelihood} of the samples, and is defined as:
\begin{equation}\label{eq:bce}
    \text{BCE}(y, \bb w^\top \phi(\bb x)) = \begin{cases}-\ln\left[\sigma\left(\bb w^\top \phi(\bb x)\right)\right], & y=1 \\ -\ln\left[1 - \sigma\left(\bb w^\top \phi(\bb x)\right)\right], & y=0 \end{cases}
\end{equation}
\Cref{fig:hilbertmaps} shows an illustration of both the hinge point feature transform and the sampling used for creating Hilbert maps.

Hilbert maps have previously been extended to Bayesian Hilbert maps, where a distribution over weights is modeled as a multivariate Gaussian \cite{senanayake2017bayesian, senanayake2018automorphing, zhi2019continuous}. There is also a multiclass variant that defines a weight \textit{matrix} with Gaussian rows described in \cite{wright2024vprism}. In this work, we adopt the Hilbert map representation, but model the distribution over weights as a collection of particles. This allows our method the capability to capture irregular, non-Gaussian posterior distributions.

\subsection{Stein Variational Gradient Descent}

\begin{figure*}[ht!]
    \centering
    \includegraphics[width=0.9\linewidth]{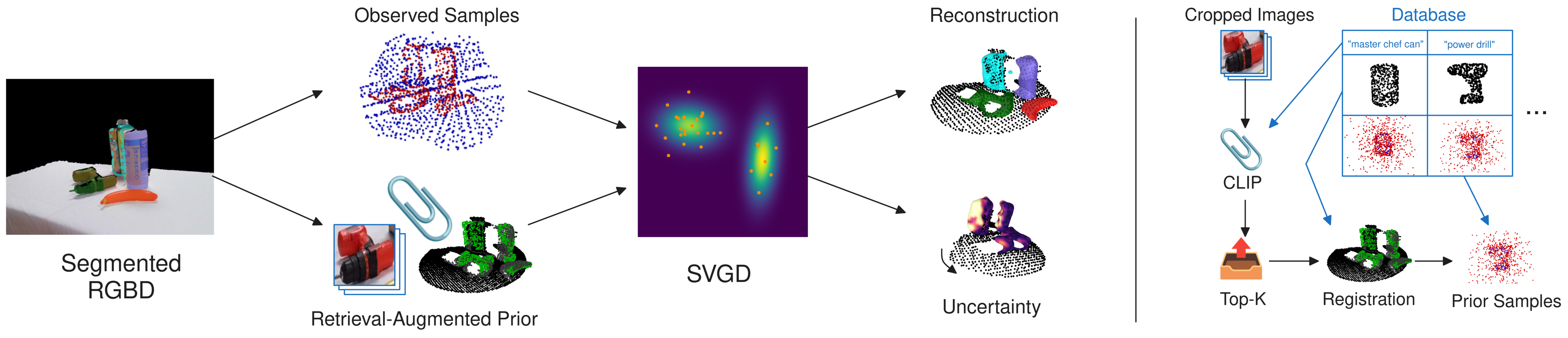}
    % \caption{Overview of BRRP method. We begin with a segmented RGBD image and (a) feed cropped images of each segment into CLIP to get object probabilities (\Cref{subsec:retrievalaug}) Then, we retrieve and (b) register the the top-\(k\) objects in the prior. This gives us a set of registered prior samples (\Cref{subsec:reconpriors}). We also (c) compute negative samples based on the observed segmented point cloud (\Cref{subsec:negativesampling}). Finally, (d) we run SVGD optimization to recover a posterior distribution over Hilbert map weights (\Cref{subsec:svgdrecon}). We can use this distribution to both reconstruct the scene as well as measure uncertainty.}
    \caption{\textbf{Left:} Overview of BRRP method. We begin with a segmented RGBD image and compute both a retrieval-augmented prior (\Cref{subsec:reconpriors} and \Cref{subsec:retrievalaug}) as well as a likelihood from negative sampling (\Cref{subsec:negativesampling}). Then, we combine both in a Bayesian manner via SVGD (\Cref{subsec:svgdrecon}). \textbf{Right:} An overview of our retrieval-augmented priors for reconstruction. We use CLIP to retrieve the top-\(k\) relevant objects in our database (\Cref{subsec:retrievalaug}), then we perform registration and retrieve the corresponding prior samples (\Cref{subsec:reconpriors}).}\label{fig:method}
    \vspace{-1.75em}
    % TODO: fix description of overview
\end{figure*}

Stein Variational Gradient Descent (SVGD) \cite{liu2016stein} is an algorithm for variational inference that closely resembles gradient descent. The general problem of variational inference is to find a distribution $q^* \in \mathcal Q$ that is close to some target distribution $p$. Usually, this takes the form of an optimization problem over the Kullback-Leibler (KL) divergence:
$$ q^* = \text{arg} \min_{q \in\mathcal Q} \mathbb{KL} (q \| p). $$
SVGD aims to iteratively transform $q$ in descent directions of the KL divergence in a $d$-dimensional reproducing kernel Hilbert space, $\mathcal H^d$. 
% We can transform our distribution $q$ to $q_{[T]}$ by applying $T(x) = x + f(x)$ with $f \in \mathcal H^d$ to the random variable corresponding to $q$. 
Because this Hilbert space is a space of functions, a descent direction requires deriving the \textit{functional gradient} of our KL divergence objective.
\begin{theorem}\label{theorem:svgd}
    From \cite{liu2016stein}. Let $T(\bb x) = \bb x + f(\bb x)$, where $f \in \mathcal H^d$ and $q_{[T]}$ is the density of random variable $\bb z = T(\bb x)$ when $\bb x \sim q$. Then
    $$ \nabla_f \mathbb{KL}(q_{[T]} \| p)|_{f=0} = - g^*_{q, p}, $$
    where $g^*_{q, p} = \mathbb E_{\bb x \sim q(\bb x)} [k(\bb x, \cdot) \nabla_\bb{x} \ln p(\bb x) + \nabla_\bb{x} k(\bb x, \cdot)]$.
\end{theorem}

In SVGD, $q$ is approximated by a set of particles $\bb x^{(0)}_1, ..., \bb x^{(0)}_P \sim q(\bb x)$. This can be used to approximate the gradient in \Cref{theorem:svgd} with $\hat{g}^*$:
\begin{equation}\label{eq:stein}
    \hat{g}^*(\bb x) = \frac{1}{P} \sum_{i=1}^{P} k(\bb x_i, \bb x) \nabla_\bb{x_i} \ln p(\bb x_i) + \nabla_\bb{x_i} k(\bb x, \cdot).
\end{equation}
The particles can then be iteratively updated according to $\hat{g}^*$ in \Cref{eq:stein} with:
$$ \bb x^{(t+1)}_i = \bb x^{(t)}_i + \epsilon \hat{g}^* (\bb x^{(t)}_i) $$
The result of these iterations is that the set of particles converges to an approximation of the target distribution $p$. Importantly, \Cref{eq:stein} only relies on the gradient of the log of $p$, which means we can perform variational inference to an unnormalized distribution. Such unnormalized distributions are common in many Bayesian inference problems.

% QUESTION: should I include a paragraph talking about interpretation of SVGD objective

\section{Retrieval-Augmented Priors}\label{sec:priors}

Retrieval-augmented generation \cite{lewis2020retrieval} was originally introduced in the context of improving language generation. The work has served as inspiration for an approach to affordance-prediction in \cite{kuang2024ram}. In our case, we draw inspiration from retrieval-augmented generation, but we use the retrieved results to improve efficiency in certain \textit{explicit} formulations for prior density functions during Bayesian inference.  

To motivate retrieval-augmented priors, consider the problem of Bayesian inference with a mixture model acting as the prior distribution. Given some data, we would like to infer a posterior distribution over hypotheses. \rebut{Defining a mixture model prior over components $C =\{c_1, \ldots, c_{N_c}\}$ gives:}
\begin{equation}\label{eq:bayes}
    P(H | D) \propto P(D | H) \textstyle{\sum_{c \in C} P(H | c)}.
\end{equation}
If our prior distribution has a lot of components, it may be inefficient to fully evaluate. This could be a serious problem for algorithms like SVGD, which requires iteratively computing the gradient of both the likelihood and prior. Inspired by \cite{lewis2020retrieval}, the insight behind retrieval-augmented priors is to determine which subset of the prior distribution components to retrieve and use given some detection result $R$. Conditioning on this detection result, we have a new posterior distribution, $P(H | D, R)$. Making an independence assumption,
$$ P(H | D, R) \propto P(D | H) \cdot \mathbb E_{c \sim P(c | R)} [P(H | c)]. $$
Comparing to \Cref{eq:bayes}, the expectation now replaces the true prior. Then, we can use a top-$k$ approximation for the expectation:
\begin{equation}\label{eq:retreivalprior}
P(H | D, R) \propto P(D | H) \textstyle{\sum_{c \in \text{topk}} P(H | c) P(c | R)}
\end{equation}
Thus, we only evaluate a subset of the prior components.

\section{The BRRP Method}\label{sec:BRRP}

Our method takes a single RGBD image and produces reconstructions for each object in the scene. We treat the problem as a Bayesian inference problem over an observation described by negative samples. We incorporate prior information on the shape of the object by leveraging retrieval-augmented priors introduced in \Cref{sec:priors}. We use the \rebut{pretrained CLIP \cite{radford2021learning} foundation model} to determine which objects to retrieve and define our object-specific priors by a registered set of pre-computed samples from the stored mesh. We then use SVGD to optimize for a set of samples over map weights. We can generate predicted reconstructions by taking the expected occupancy over our weights for a given location. \Cref{fig:method} shows a visual overview of our method.

In \Cref{subsec:reconpriors}, we explain how we leverage pre-existing mesh assets to create a prior that is robust to different poses and scales. Then, in \Cref{subsec:retrievalaug}, we explain how we utilize the retrieval-augmented priors paradigm to retrieve relevant objects in the prior. The specific negative sampling is explained in \Cref{subsec:negativesampling}. Then we give the specific SVGD objective used in \Cref{subsec:svgdrecon}.

\subsection{Negative Samples as Reconstruction Priors}\label{subsec:reconpriors}

We want to leverage existing mesh assets as our priors during Bayesian reconstruction. We define our prior as a mixture \rebut{model with different objects serving as each component}, $c_1, ..., c_{N_c}$. Because there is not a direct way to convert a mesh into a Hilbert map \rebut{prior}, we instead \textit{sample} points $\tilde{\bb x}_{c, 1}, ..., \tilde{\bb x}_{c, Q} \in \mathbb R^3$ around each object $c$'s mesh. We refer to these samples as the \textit{prior samples}. We give them labels $\tilde y_{c, 1}, ..., \tilde y_{c, Q} \in \{-1, 1\}$ determined by whether they are outside or inside the mesh. Then we simply define our prior using this data and a Gaussian prior over weight norm:
\begin{flalign}\label{eq:prior}
    & \hat P(\bb w | c) :=  P(\{\tilde y_{c, i}\} | \{\tilde {\bb x}_{c, i}\}, \bb w) P(\bb w) & \\
    &\hspace{8mm}\propto \exp(\lambda \|w\|^2) \prod_{i=1}^Q \exp\left(\text{BCE}(\tilde y_{c, i}, \bb w^\top \phi(\tilde{\bb x}_{c, i})\right).& 
    \label{eq:prior2}
\end{flalign}
%where $\text{BCE}$ is the same as in \Cref{eq:bce}. 

In order to enforce pose-invariance \rebut{for each prior component}, we first register a small stored point cloud of the \rebut{respective} object to the observed points and then transform the prior samples to this reference frame. In practice we use RANSAC \cite{fischler1981random} and the FPFH features from \cite{rusu2009fast} to perform registration. In order to also have scale invariance, we do a linear scan over 10 different scales and select the the scale that resulted in the most inlier pairs from the registration.

\subsection{Retrieval-Augmented Priors for Hilbert Maps}\label{subsec:retrievalaug}

Because it would be inefficient to register all meshes that are part of the prior mixture model, we propose using the retrieval-augmented prior approach introduced in \Cref{sec:priors}. In order to determine which objects to use, we need to compute $P(c | R)$ from \Cref{eq:retreivalprior}. In our case, we use CLIP \cite{radford2021learning} as a zero-shot classifier for our different objects. For each object in our prior, we store a small textual description of the object. These descriptions are then used as classes for CLIP to classify each segmented object. In order to make sure CLIP knows which object we are targeting, we crop the RGB image to fit the predicted segmentation of each object. \rebut{ One could consider these cropped images to be the $R$ in our $P(c|R)$ formulation.} We feed the cropped images as input into CLIP. \rebut{The resulting distribution over objects predicted by CLIP becomes our $P(c | R)$ for each component.} 

Once we have the probability of each object, we retrieve and register the stored point clouds of the top-\(k\) objects. After registration, we retrieve the prior samples corresponding to these objects to define our \rebut{top-\(k\) component prior approximation} according to \Cref{eq:prior}.
\subsection{Negative Sampling}\label{subsec:negativesampling}
% TODO: fix this section. I think it may be incorrect.
We adopt the negative sampling method introduced in \cite{wright2024vprism}. The negative sampling method makes the assumption that all objects are lying on or above a planar surface. We begin by labeling the points segmented to each object as occupied for that object. Next, we perform stratified sampling along each camera ray near each object to recover a set of negatively sampled points, labeled as unoccupied. Then, we use RANSAC over points not segmented to any object to recover the flat surface all objects are resting on. This plane is used to randomly sample points in a sphere underneath each object that are near the object. We also label these points as occupied. Finally, we use grid subsampling from \cite{thomas2019learning} to reduce the number of points and increase uniformity of sampled points. We refer to these points and labels as \textit{observed samples} and denote them as $\{\bb x_i\}_{i \in [S]}, \{y_i\}_{i \in [S]}$. The entire negative sampling process can be easily parallelized for efficient computation.

\begin{figure}
    \centering
    \includegraphics[width=.8\linewidth]{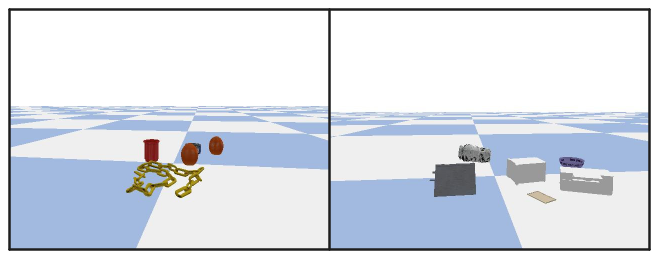}
    \caption{Sample images of procedurally generated scenes used to evaluate BRRP. \textbf{Left:} a YCB scene. \textbf{Right:} a ShapeNet scene.}
    \label{fig:proceduralscenes}
\end{figure}

\subsection{SVGD Reconstruction}\label{subsec:svgdrecon}

Once we have retrieved our prior samples and computed our observed samples, we can perform optimization-based reconstruction with SVGD. Given both sets of samples and our prior definition from \Cref{eq:prior}, we have the following posterior distribution,
$$ P(\{y_{c, i}\} | \{{\bb x}_{c, i}\}, \bb w) P(\bb w) \sum_{c \in \text{topk}} P(c | R) P(\{\tilde y_{c, i}\} | \{\tilde {\bb x}_{c, i}\}, \bb w). $$
Taking the log and applying \Cref{eq:prior2} gives the objective:
%\begin{align*} 
%    &\ell = \sum_{i = 1}^{S} \text{BCE}(y_i, \bb w^\top \phi(\bb x_i)) \\ 
%    &+ \sum_{c \in \text{topk}} P(c|R) \ln \left[\sum_{i = 1}^{Q} \exp\left(\text{BCE}(\tilde y_{c, i}, \bb w^\top \phi(\tilde{\bb x}_{c, i}))\right)\right] \\
%    &+ \lambda \|\bb w\|^2 + \text{const.}
%\end{align*}
\begin{align} 
    &\ell = \frac{\lambda_\text{\rebut{1}}}{S} \sum_{i = 1}^{S} \text{BCE}(y_i, \bb w^\top \phi(\bb x_i))\label{eq:obj1} \\ 
    &+ \frac{\lambda_2}{K} \sum_{c \in \text{topk}} P(c | R) \ln \left[ \frac{1}{Q}\sum_{i = 1}^{Q} \exp\left(\text{BCE}(\tilde y_{c, i}, \bb w^\top \phi(\tilde{\bb x}_{c, i}))\right)\right]\label{eq:obj2} \\
    &+ \lambda_\text{3} \|\bb w\|^2\label{eq:obj3},
\end{align}
\rebut{where $K$ is the number of objects retrieved for the prior. For practical reasons, we introduce multipliers to each term as hyperparameters during optimization and use means instead of sums for the first two terms.
}
This objective is used as the log of the target distribution, $\ln P(\bb w)$, in \Cref{eq:stein}, where we also adopt the original median kernel suggested in~\cite{liu2016stein}. We also opt to use SVGD in a stochastic manner, where both the observed samples and query samples are mini-batched.

From a non-probabilistic standpoint, one can interpret \Cref{eq:obj1} as the likelihood of the observed data, \Cref{eq:obj2} as the object shape prior, and \Cref{eq:obj3} as regularization. 

\section{Experiments}\label{sec:experiments}

In this section, we aim to experimentally validate the following claims: (1) BRRP is more \textit{robust} than deep learning methods; (2) BRRP is more \textit{accurate} than methods that use uninformative priors; (3) BRRP can capture principled uncertainty. (4) BRRP's robustness improves the downstream success of grasping. First, we provide details on the experiments, such as the baselines and metrics in \Cref{subsec:details}, then we show then show the results and analyses in \Cref{subsec:results}.

\begin{figure}
    \centering
    \raisebox{-\height}{\includegraphics[width=0.20\textwidth]{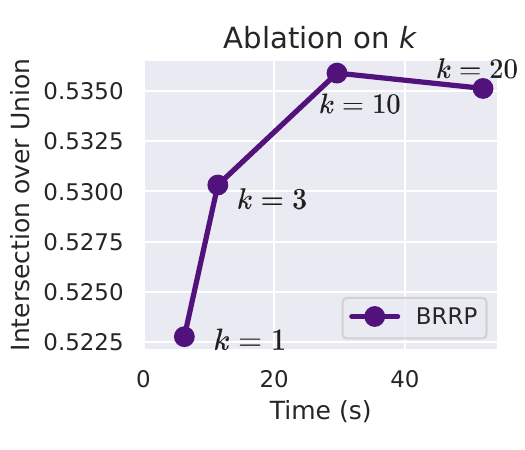}}
    \hfill
    \raisebox{\dimexpr-\height-0.11cm\relax}{\includegraphics[width=0.282\textwidth]{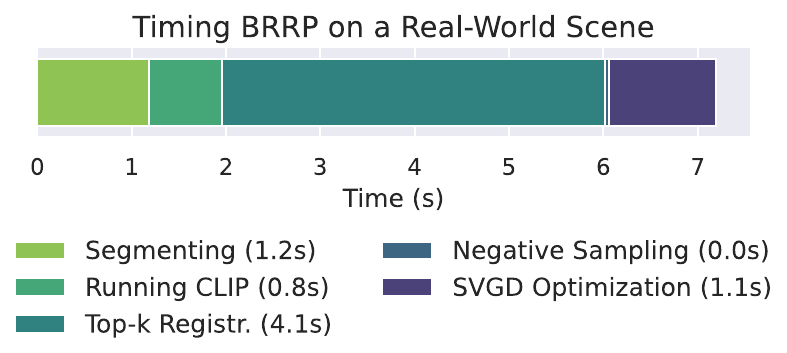}}
    \caption{\rebut{Results pertaining to the speed of BRRP on an NVIDIA RTX 4070 GPU. \textbf{Left:} Ablation on $k$ for YCB scenes highlights the accuracy-speed tradeoff in retrieval-augmented priors. We use $k=3$ for all other experiments. \textbf{Right:} Timing breakdown of BRRP on a real-world scene with 4 objects.}}
    \label{fig:ablation_and_speed}
\end{figure}

\subsection{Experimental Details}\label{subsec:details}

\textbf{BRRP Implementation:} We use a set of 50 objects from the YCB dataset \cite{calli2015ycb} to act as the prior for our experiments with BRRP. We implement the method in PyTorch. \rebut{We also display results for the speed of running BRRP on an NVIDIA RTX 4070 GPU in \Cref{fig:ablation_and_speed}. We use $k=3$ for all experiments, but note that our method's performance is fairly robust to the $k$ parameter (\Cref{fig:ablation_and_speed}).}

\textbf{Baselines:} We compare our work against two main baselines, V-PRISM \cite{wright2024vprism} and a version of PointSDF \cite{van2020learning} that predicts occupancy and is trained on ShapeNet \cite{chang2015shapenet} scenes. V-PRISM is a probabilistic mapping method that uses an uninformative prior. This means that it is robust to novel objects, but doesn't accurately reconstruct object backsides. We refer to this baseline as \textbf{V-PRISM}. In contrast, PointSDF is a learning-based method. This means it can leverage prior information from mesh datasets to accurately reconstruct the backside, but can suffer in performance under significant distributional shift. \rebut{We train two versions of PointSDF. One is trained on scenes containing a wide variety of Shapenet objects, and another trained only on scenes containing the 50 YCB objects in our BRRP prior}. We refer to this baseline as \textbf{PointSDF}\rebut{, and generally use the ShapeNet version when the training dataset is not mentioned}. When  reconstructing meshes with PointSDF, a level set of $\tau = 0.3$ is used.

\begin{table}
    \centering
    \adjustbox{width=\linewidth}{
    \begin{tabular}{c c c}
        \toprule
        \textbf{Method} & \textbf{ShapeNet Scenes (IoU \(\uparrow\))} & \textbf{YCB Scenes (IoU \(\uparrow\))}
        \\
         \midrule
         V-PRISM \cite{wright2024vprism} & 0.3092 & 0.5003
         \\[2pt]
         PointSDF (ShapeNet) \cite{van2020learning} & \textbf{0.3600} & 0.4601
         \\[2pt]
         \rebut{PointSDF (YCB)\cite{van2020learning}} &  \rebut{0.2426} & \rebut{0.5088} 
         \\
         \midrule
         BRRP (ours) & 0.3124 & \textbf{0.5277} 
         \\
         \bottomrule \\
    \end{tabular}
    }
    \caption{Intersection over union (IoU) on procedurally generated scenes from \rebut{two} different mesh datasets. BRRP uses a YCB prior and \rebut{there are two different PointSDF versions, trained on ShapeNet and YCB scenes respectively. BRRP outperforms V-PRISM, which has an uniformative prior, whereas BRRP is more robust to distributional shift than PointSDF.}}
    \label{table:iou}
\end{table}

\begin{figure}
    \centering
    \includegraphics[width=0.95\linewidth]{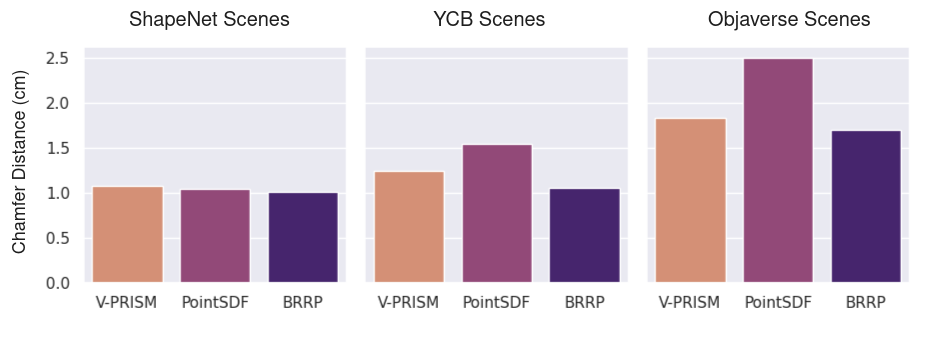}
    \caption{Chamfer distances (lower is better) for various methods across the procedurally generated scenes. Values are reported in centimeters. BRRP has the lowest chamfer distance on each dataset.}
    \label{fig:chamfer}
\end{figure}

\begin{figure}[ht]
    \centering
    \includegraphics[width=1\linewidth]{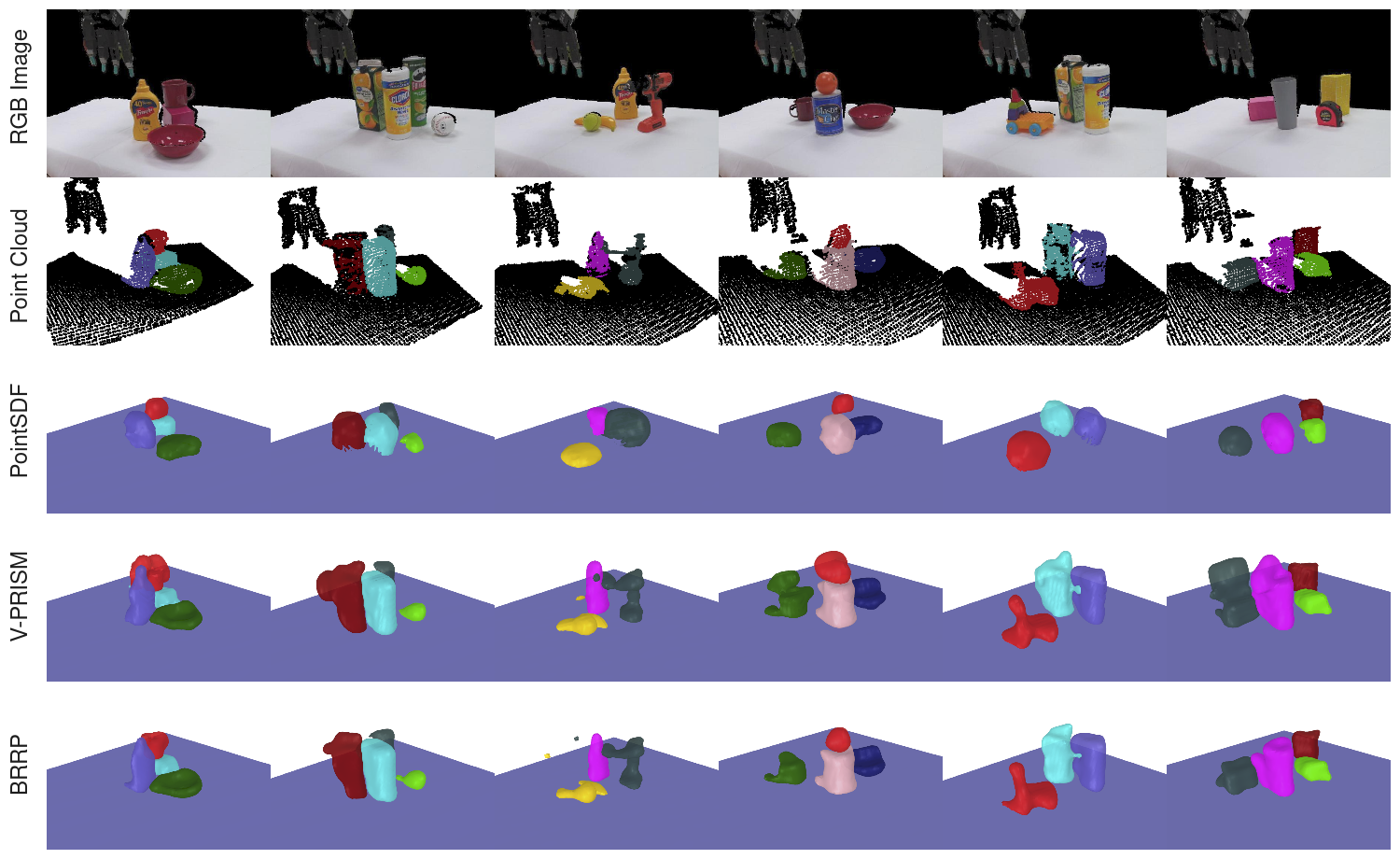}
    \caption{Qualitative comparison of BRRP and our baselines. PointSDF tends to predict a spherical shape for many non-spherical objects. V-PRISM can sometimes predict occupancy in portions of the scene that are not occupied. Our method is more robust and can more accurately reconstruct the scenes.}
    \label{fig:qualitative1}
\end{figure}

\textbf{Procedurally Generated Scenes:} We use the generated scenes from \cite{wright2024vprism} to evaluate our method. These scenes are constructed with objects from ShapeNet \cite{chang2015shapenet}, YCB \cite{calli2015ycb}, and Objaverse \cite{deitke2023objaverse} datasets. There are 100 multi-object scenes for each mesh dataset. Each scene contains up to 10 objects. Some meshes in the Objaverse and ShapeNet scenes did not have correctly rendered textures and were instead rendered as plain white objects.  \Cref{fig:proceduralscenes} contains two example images of these procedurally generated scenes. \rebut{We use the ground-truth masks for experiments on procedural scenes.} We also conduct an experiment on robustness where we perturb the masks of the ShapeNet scenes by two pixels and evaluate reconstructions.

We evaluate performance on the procedurally generated scenes with two metrics: intersection over union (IoU) and chamfer distance. We refer readers to \cite{wright2024vprism} for further explanation of these metrics.

\textbf{Real World Scenes:} In order to showcase robustness to real-world noise, we evaluate on real world scenes collected with a Kinect depth camera. In order to obtain instance segmentations, we use Grounded SAM \cite{ren2024grounded} along with some depth filters. \rebut{ We include timing of performing this segmentation in the real-world timing breakdown in \Cref{fig:ablation_and_speed}.} We evaluate on these real world scenes qualitatively with images of scene reconstruction and visualizing surface uncertainty.

\begin{figure}
    \centering
    \includegraphics[width=0.9\linewidth, trim=5cm 5cm 10cm 15cm, clip]{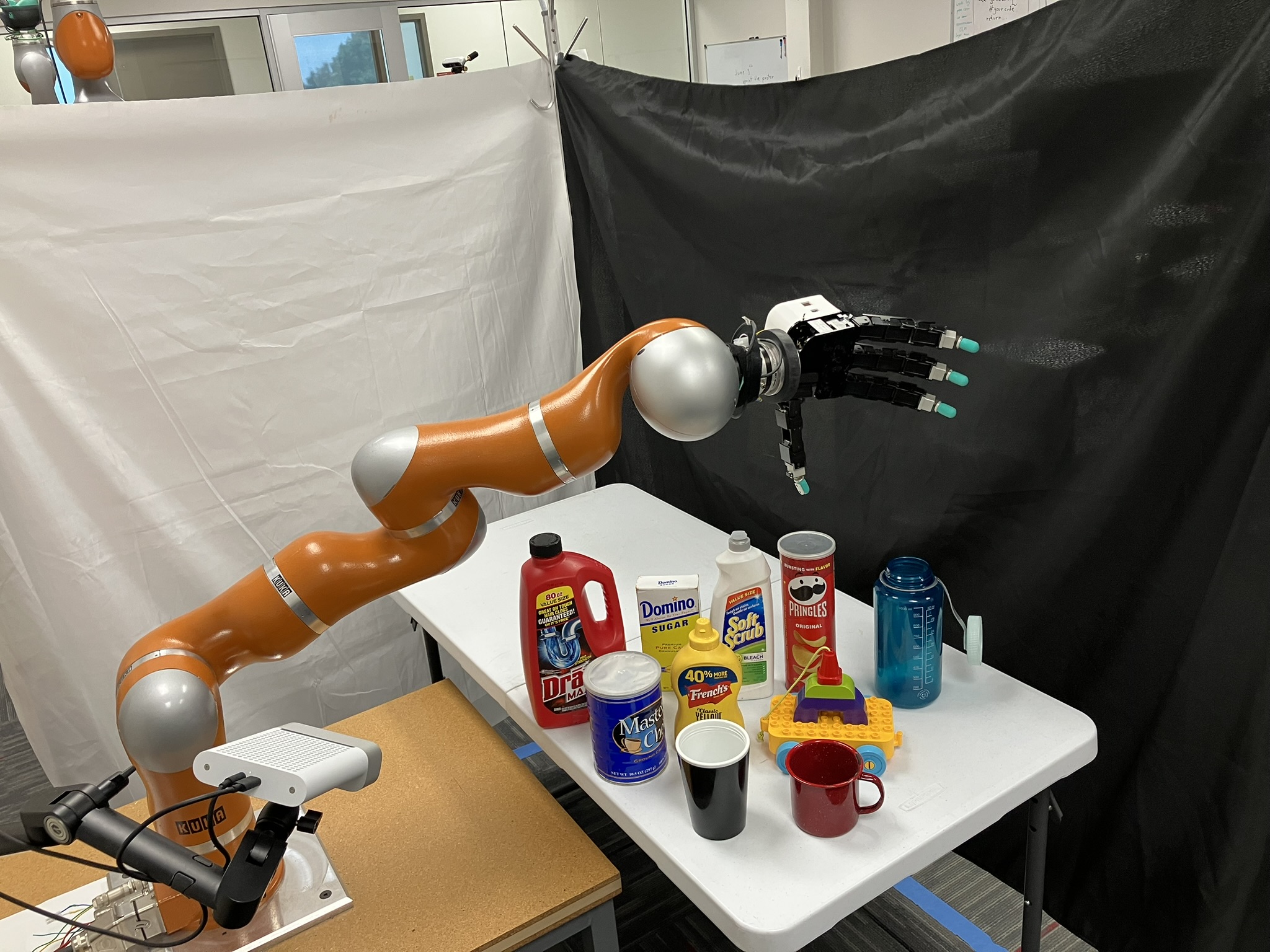}
    \caption{
    %\textbf{Left:}
    Objects and the robot used to assess the impact of our method on dexterous grasping. %\textbf{Right:} the results for 50 grasps.
    }
    \label{fig:grasp-experiments}
\end{figure}

\textbf{Grasping Experiment:} To understand the downstream impact of our method, we assess its effectiveness on the problem of dexterous grasping from a fixed RGB-D camera. We place an object on the table, segment it, reconstruct its mesh, plan a grasp trajectory, and execute it open-loop. \Cref{fig:grasp-experiments} shows the 10 objects used.

For segmentation, we use Grounded SAM~\cite{ren2024grounded} with the prompt \textit{object that can be picked up with one hand}. Meshes are reconstructed with marching cubes on either BRRP or PointSDF reconstructions. We implement a grasp planner~\cite{martin-arxiv-fpte} which takes a segmented point cloud as input and moves the robot to a pregrasp configuration that is collision-free according to the reconstruction while also avoiding obstacles according to the scene reconstruction. The robot then closes the hand and lifts the arm. An attempt is successful if the object is lifted. We use KUKA LBR4 with Allegro hand in the experiment. We place each of the 10 objects across 5 different poses on the table, resulting in 50 executed grasp trajectories for each method. \blue{We select stable poses where objects are reachable by the robot, visible to the camera, and remain stationary, while ensuring diversity in positions and orientations.} We report the grasp success rate per method across all 50 trajectories.

\subsection{Results \rebut{\& Discussion}}\label{subsec:results}

In \Cref{table:iou}, we display the IoU results from procedurally generated scenes. The chamfer distances for the procedurally generated scenes is shown in \Cref{fig:chamfer}. The qualitative reconstructions on real world scenes can be seen in \Cref{fig:qualitative1}.

\noindent\textbf{Insight 1:} \textit{BRRP is more accurate than a method with an uninformative prior.}

As showcased in \Cref{table:iou}, BRRP outperforms V-PRISM on \rebut{both sets of} procedurally generated scenes. It has the highest IoU improvement from V-PRISM on the YCB scenes. A similar pattern can be seen in the chamfer distances in \Cref{fig:chamfer}, where BRRP consistently outperforms V-PRISM, with the biggest gap of 0.19 (cm) occuring on the YCB scenes. This makes sense because BRRP uses a subset of the YCB objects as its prior.

We can see this improvement qualitatively in \Cref{fig:qualitative1}. While BRRP and V-PRISM are comparable for most objects, there exist certain objects that V-PRISM predicts to occupy a large portion of space that the object \textit{doesn't} occupy. BRRP is able to more accurately reconstruct these objects. The clearest example of this is the dark green object in the right-most scene in \Cref{fig:qualitative1}. Both of these quantitative and qualitative results suggest BRRP is generally more accurate than V-PRISM. It is the most accurate when evaluated on objects in its prior distribution.

\begin{figure}
    \centering
    \includegraphics[width=1\linewidth]{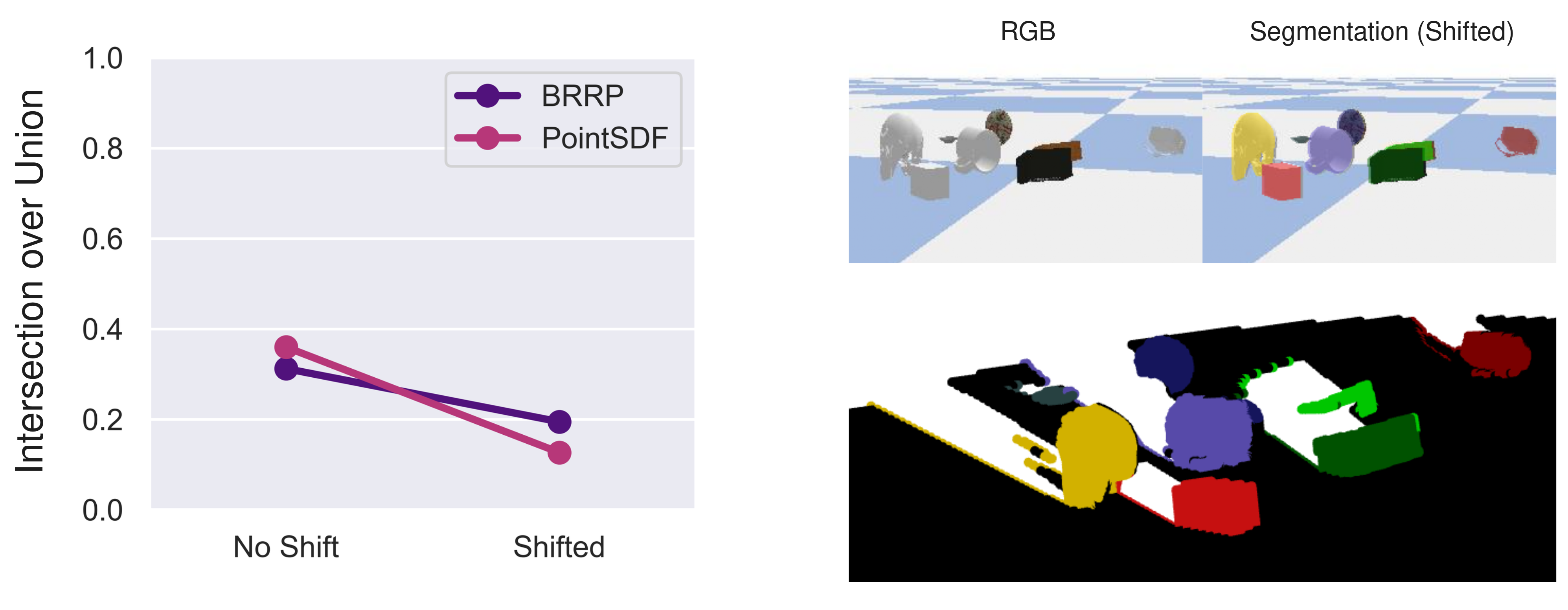}
    \caption{(\textbf{left}) IoU of BRRP and PointSDF on ShapeNet scenes with and without shifted segmentations. Our method is more robust to segmentation shifts. (\textbf{right}) An example of a scene and the corresponding point cloud with shifted segmentation.}
    \label{fig:shift}
\end{figure}

\noindent\textbf{Insight 2:} \textit{BRRP is more robust than a deep learning method.}

While \rebut{the ShapeNet version of} PointSDF outperformed BRRP on the ShapeNet scenes on IoU (\Cref{table:iou}), BRRP had a lower Chamfer distance (\Cref{fig:chamfer}). \rebut{BRRP also outperformed the YCB version of PointSDF on both YCB and ShapeNet (\Cref{table:iou}).} \rebut{Generally, BRRP performed significantly better than each PointSDF version on mesh datasets the specific version was not trained on.} 
When measuring chamfer distance, BRRP outperformed \rebut{the ShapeNet version of} PointSDF on all datasets as shown in \Cref{fig:chamfer}, \rebut{including the Objaverse dataset, where objects were out-of-distribution for both methods}. These results suggests BRRP is more robust to different object distributions than PointSDF. Next, we evaluate robustness to slightly incorrect \rebut{image} segmentations. We take the procedurally generated ShapeNet scenes and shift the segmentation over by 2 pixels. In \Cref{fig:shift}, we compare the IoU of BRRP and \rebut{ShapeNet version of} PointSDF on the scenes with and without the shift. Our method performs better on the shifted scenes compared to PointSDF.

On the real-world scenes in \Cref{fig:qualitative1}, BRRP is qualitatively more robust than PointSDF. PointSDF struggles with the noise inherent in real world scenes as well as novel objects. It tends to predict a spherical object on many non-spherical objects in the real world scenes. Our method is better able to reconstruct these scenes, including the objects that are out-of-distribution for its prior. These results suggest BRRP is more robust than PointSDF. Even though by one metric the ShapeNet-trained PointSDF outperforms BRRP on ShapeNet scenes, when the scenes are perturbed, BRRP performs better.

\begin{figure}
    \centering
    \includegraphics[width=1.0\linewidth]{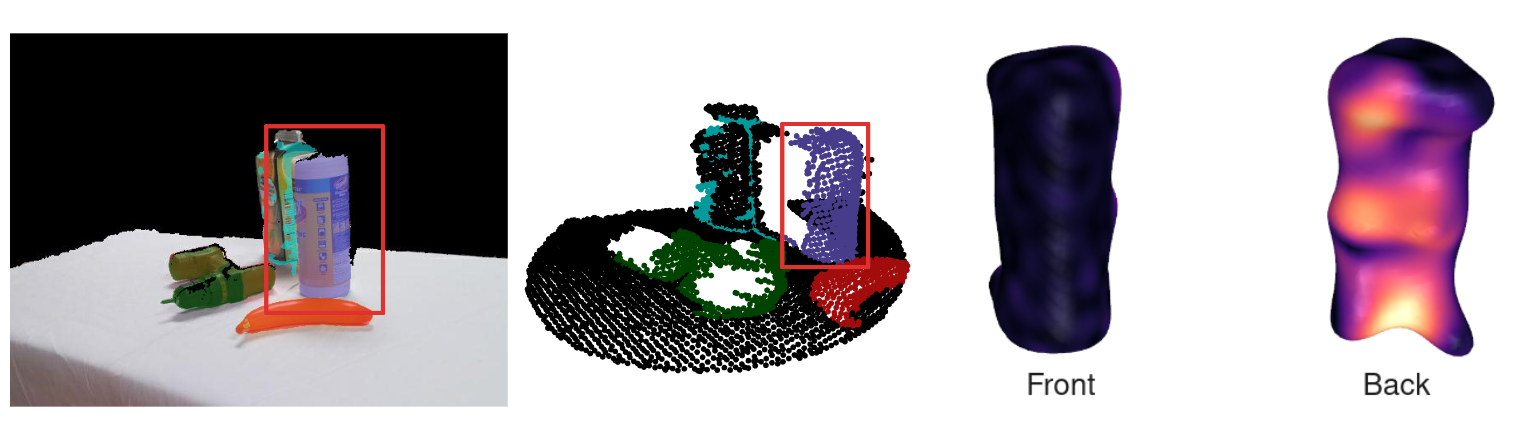}
    \caption{Visualization of a cylindrical Clorox container surface uncertainty from BRRP. Lighter areas correspond to higher uncertainty about the shape. We observe that the occluded back-side of the container has high uncertainty. }
    \label{fig:uncertainty}
    \vspace{-1.25em}
\end{figure}

\begin{figure}
    \centering
    \includegraphics[width=0.245\linewidth]{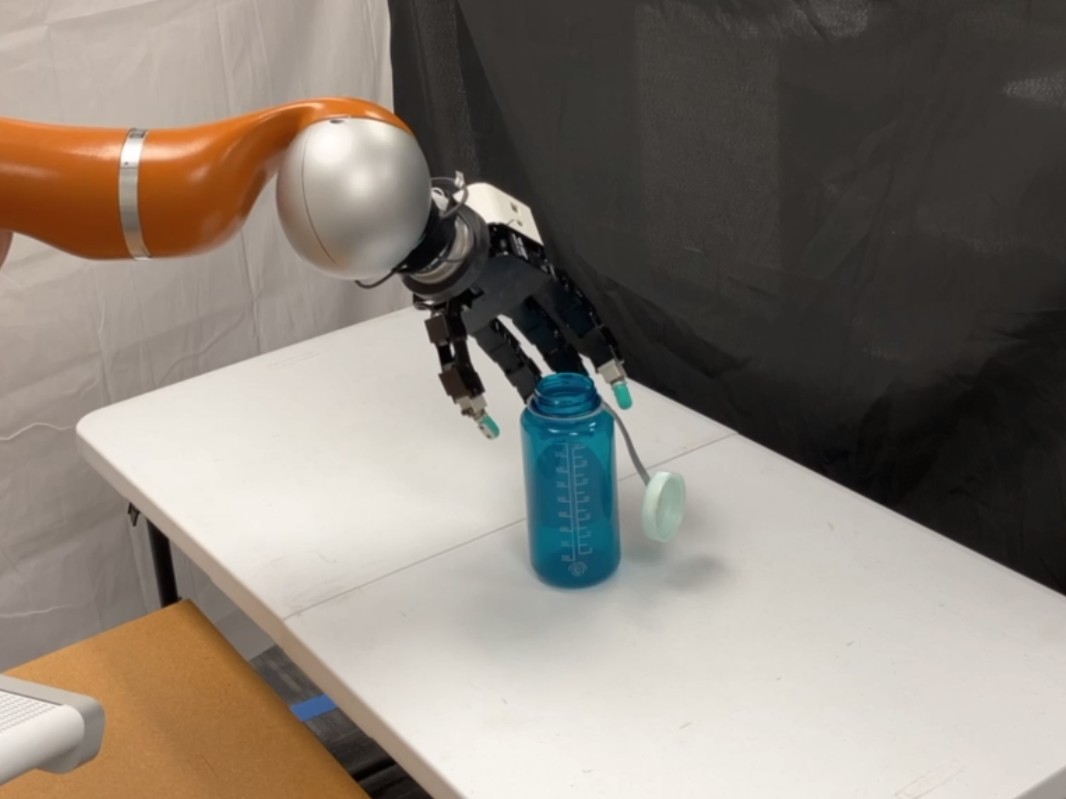}%
    \includegraphics[width=0.245\linewidth]{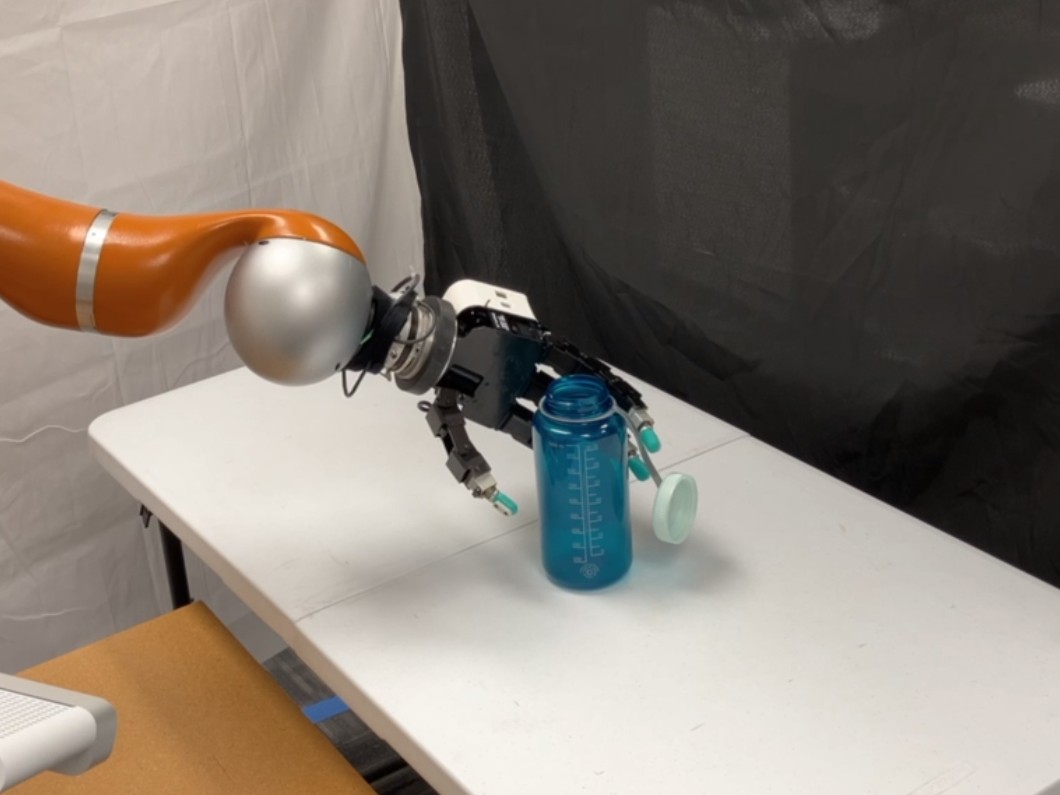}%
    \includegraphics[width=0.245\linewidth]{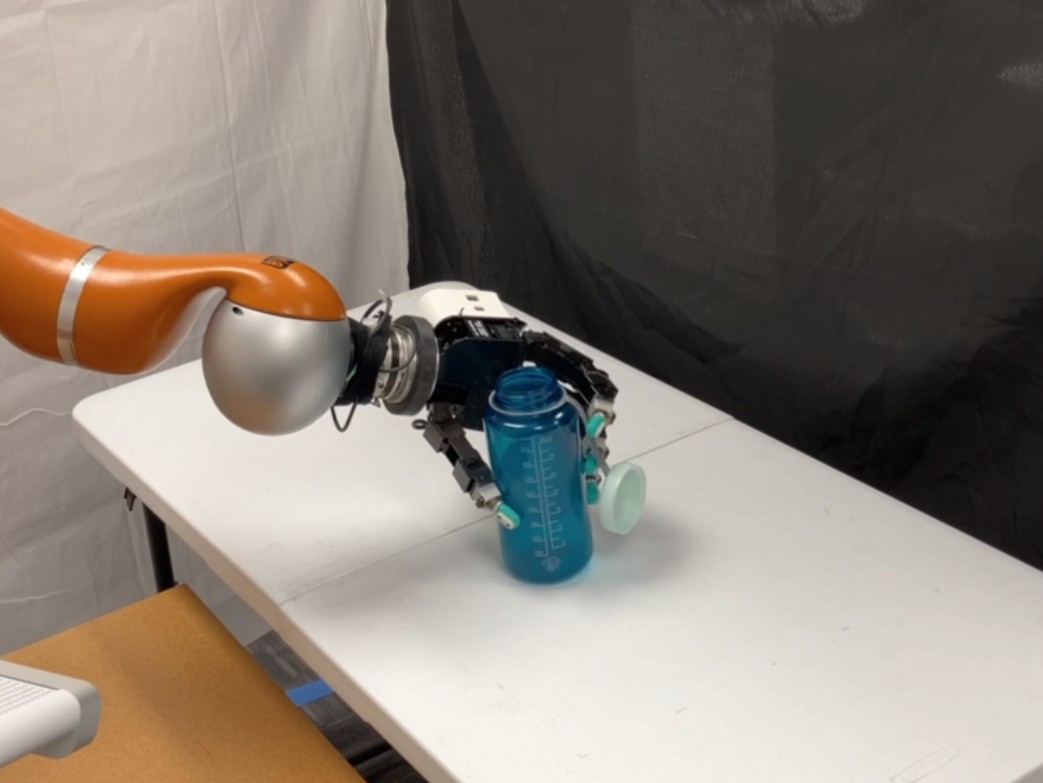}%
    \includegraphics[width=0.245\linewidth]{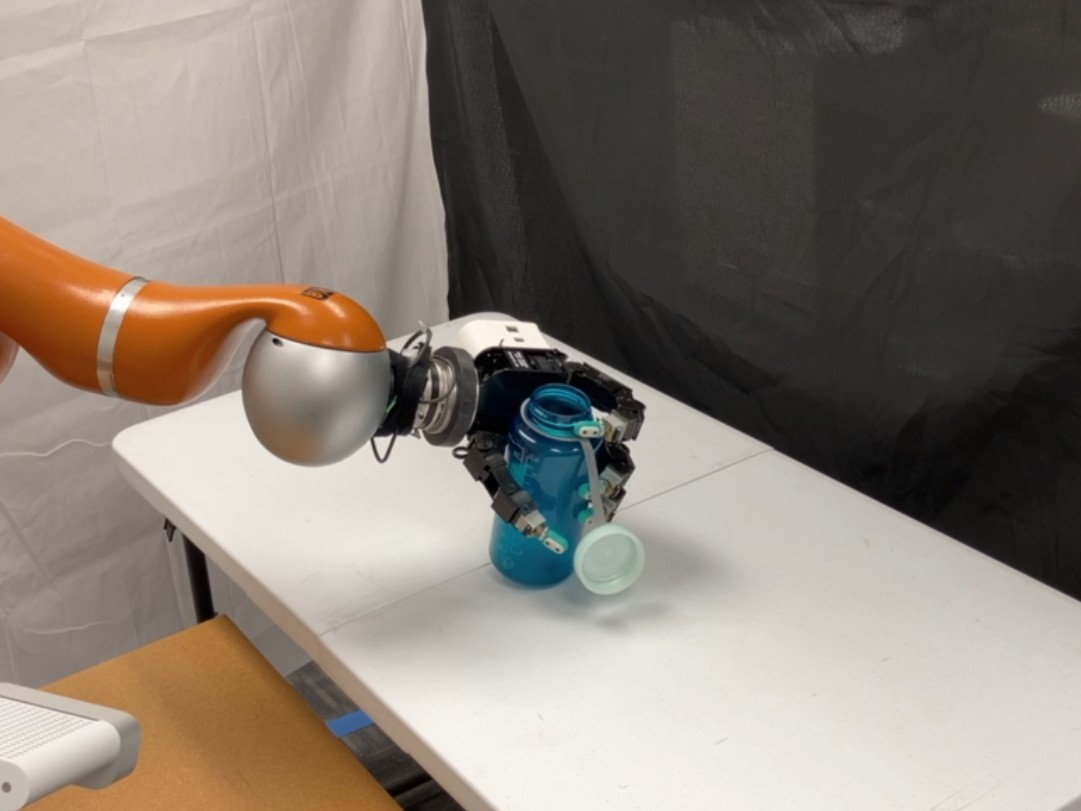}%
    \caption{BRRP builds precise representations, which are critical for downstream dexterous manipulation. Here, the water \rebut{(out-of-distribution)} is particularly large, and the dexterous hand needs to arrive at a pre-grasp pose that is very close to the surface of the bottle, all the while requiring the fingers to avoid colliding.}
    \label{fig:grasp}
\end{figure}

\noindent\textbf{Insight 3:} \textit{BRRP can capture principled uncertainty about object shape.}

\Cref{fig:uncertainty} shows a qualitative example of uncertainty from BRRP. We measure the uncertainty by taking the variance of logits over weight particles. Our method predicts the highest uncertainty in areas of the surface that are occluded. This suggests that we can utilize surface uncertainty from BRRP in a similar way to how GPIS surface uncertainty is utilized in many grasping applications.

\noindent\textbf{Insight 4:} \textit{BRRP's robust reconstructions improve the downstream success of dexterous grasping.}

The success rate of BRRP and the PointSDF baseline is 64\% and 40\% respectively on the downstream grasping experiment. The difference in success rates stems from the modeling of the obstacles, consisting of the reconstructed object mesh and the table (fixed). The collision modeling directly impacts how close the robot can get to the object before executing the grasp. This directly affects whether the object will get knocked over or bumped into as the gripper moves to the pre-grasp configuration. BRRP exhibits higher accuracy and robustness in reconstruction, which creates an over 50\% increase in grasp success rates. \Cref{fig:grasp} illustrates an example and highlights the impact of BRRP in significantly improving downstream grasp success. \rebut{While the difference in BRRP and PointSDF grasp success can be easily attributed to reconstruction quality, we hypothesize the 36\% of grasp failures with BRRP stem from a mixture of reconstruction, grasp selection, and motion planning inaccuracies.}

\rebut{\textbf{Limitations:} As shown in \Cref{fig:ablation_and_speed}, the registration portion of our method grows with the number of objects, which can be resolved by fast learning-based registration methods. While our method is robust to viewpoint, occlusion, and scale; it is still limited by the quality of segmentation, depth accuracy, and its prior database. Advances in  segmentation methods will further improve BRRP performance. While BRRP can robustly handle OOD data, the reconstruction and uncertainty remain marginally biased towards the in-distribution dataset, which is common for data-driven approaches. This can be seen in \Cref{fig:fig1} and \Cref{fig:qualitative1}, and can be mitigated with a larger, more diverse prior database. }

\vspace{-0.25em}
\section{Conclusion}\label{sec:conclusion}
\vspace{-0.25em}
We introduced the concept of retrieval-augmented priors, where we retrieve relevant components of a prior distribution during Bayesian inference. We also introduced a novel Bayesian method for scene reconstruction that uses \textit{informative priors}. Our method, BRRP, leveraged existing mesh datasets to build its prior and probabilistically reconstructed the scene leveraging these priors in a retrieval-augmented manner. We showed our method was more robust than a deep learning method as well as more accurate than a mapping method. We also showed a qualitative example of recovering principled uncertainty from our method. Finally, we demonstrated that our robust reconstructions improve downstream performance, where BRRP reconstructions enable precise collision avoidance as the robot moves to execute grasps.

\bibliographystyle{IEEEtran}
\bibliography{refs} % Entries are in the "refs.bib" file

% I know RA-L doesn't allow an appendix, but I figured I could just put hyperparameters here.
% \appendix

% \subsection{BRRP Hyperparameters in Experiments}

% On both procedural and real-world experiments reported in \Cref{sec:experiments}, we use the hyperparameters reported in the following table for BRRP:

% \begin{table}[h]
%     \centering
%     \begin{tabular}{c c}
%         \toprule
%         \textbf{Parameter} & \textbf{Value} \\
%         \midrule
%         $k$ & 3\\
%         $\lambda_1$ & $50.0$ \\
%         $\lambda_2$ & $1.0$ \\
%         $\lambda_3$ & $1 \times 10^{-6}$ \\
%         SVGD batch size (obs) & 2048 \\
%         SVGD batch size (prior) & 256 \\
%         SVGD optimization algo. & Adam \\
%         SVGD learning rate & 0.1 \\
%         SVGD num. epochs & 10 \\
%         SVGD num. particles & 8 \\
%         SVGD kernel & median Gauss. \\
%         num. hinge points & $12^3$ \\
%         \bottomrule
%     \end{tabular}
%     \caption{BRRP Hyperparameters for experiments}
%     \label{tab:placeholder}
% \end{table}

\end{document}